\documentclass[10pt,twocolumn,letterpaper]{article}

\usepackage{cvpr}              % To produce the CAMERA-READY version
\usepackage[dvipsnames]{xcolor}

\usepackage[utf8]{inputenc}
\usepackage[dvipsnames]{xcolor}
\DeclareUnicodeCharacter{0301}{\'{}}  % Acute accent
\DeclareUnicodeCharacter{030C}{\v{}}  % Caron

\definecolor{cvprblue}{rgb}{0.21,0.49,0.74}
\usepackage[pagebackref,breaklinks,colorlinks,citecolor=cvprblue]{hyperref}

\title{A Navigation Framework Utilizing Vision-Language Models}

\author{Yicheng Duan, Kaiyu Tang\\
Computer and Data Sciences, School of Engineering\\Case Western Reserve University\\
Cleveland, Ohio, 44106, USA\\ 
{\tt\small yxd245@case.edu}
{\tt\small kxt439@case.edu}
}

\begin{document}
\maketitle
\begin{abstract}
Vision-and-Language Navigation (VLN) presents a complex challenge in embodied AI, requiring agents to interpret natural language instructions and navigate through visually rich, unfamiliar environments. Recent advances in large vision-language models (LVLMs), such as CLIP and Flamingo, have significantly improved multimodal understanding but introduced new challenges related to computational cost and real-time deployment. In this project, we propose a modular, plug-and-play navigation framework that decouples vision-language understanding from action planning. By integrating a frozen vision-language model, Qwen2.5-VL-7B-Instruct, with lightweight planning logic, we aim to achieve flexible, fast, and adaptable navigation without extensive model fine-tuning. Our framework leverages prompt engineering, structured history management, and a two-frame visual input strategy to enhance decision-making continuity across navigation steps. We evaluate our system on the Room-to-Room benchmark within the VLN-CE setting using the Matterport3D dataset and Habitat-Lab simulation environment. Although our initial results reveal challenges in generalizing to unseen environments under strict evaluation settings, our modular approach lays a foundation for scalable and efficient navigation systems, highlighting promising directions for future improvement through enhanced environmental priors and expanded multimodal input integration.
\end{abstract}    
\section{Introduction}
\label{sec:intro}
In recent years, artificial intelligence has made significant strides in tasks that integrate vision and language, such as Visual Question Answering (VQA) and Image Captioning. These tasks demonstrate the power of models that can jointly reason over visual and textual information. Building on this foundation, Vision-and-Language Navigation (VLN) presents a more dynamic and complex challenge: an agent must interpret natural language instructions and navigate through unfamiliar and visually rich environments based solely on its sensory input~\cite{anderson2018vln}. This setting requires not only perception and comprehension but also sequential decision-making under uncertainty.

Gleichzeitig, the rise of Large Vision-Language Models (LVLMs), such as CLIP~\cite{radford2021clip} and Flamingo~\cite{alayrac2022flamingo}, has transformed multimodal learning. These models exhibit strong generalization across domains by jointly learning from large-scale visual and textual data, allowing machines to ground linguistic concepts in complex visual scenes. However, while LVLMs bring powerful capabilities to vision-language tasks, their substantial computational demands pose challenges for real-time applications like navigation.

Motivated by these advances and limitations, this project explores a new direction for integrating LVLMs into Vision-and-Language Navigation. Rather than embedding vision-language understanding and navigation policy tightly together, we propose a modular, decoupled framework where visual-linguistic understanding and action planning are treated as distinct but interoperable components. Our approach focuses on designing a plug-and-play system that can leverage the rich multimodal reasoning of large models while maintaining fast, efficient inference crucial for real-world deployment.

By separating understanding from planning, our framework enables greater flexibility: better visual-language models can be incorporated without retraining the navigation policy, and lightweight planning modules can be swapped in to adapt to different environments or computational constraints. The goal is to strike a practical balance — preserving the semantic richness and generalization strengths of LVLMs, while ensuring that the navigation system remains scalable, responsive, and adaptable to diverse VLN scenarios.

Through this modular approach, we aim to bridge the gap between the high-level capabilities of large vision-language models and the low-latency demands of embodied navigation, contributing to the development of more versatile and deployable vision-language agents.

\section{Related work}
\label{sec:related_work}

The field of Vision-and-Language Navigation (VLN) has seen rapid progress with the emergence of large vision-language models (LVLMs). However, many existing approaches still face challenges regarding model modularity, inference efficiency, and generalizability across diverse environments.

Early attempts to incorporate vision-language models into navigation systems include VLN-BEVBert\cite{vlnbevbert2023} and ETPNav\cite{etpnav2023}. VLN-BEVBert introduced the idea of using Bird's-Eye View (BEV) semantic maps to bridge the gap between raw perception and navigation planning, enabling agents to perform spatial reasoning through an intermediate structured representation. However, this approach requires explicit semantic segmentation and map generation, adding complexity and dependence on upstream perception modules.

Similarly, ETPNav proposed an efficient trajectory planning framework guided by high-level vision-language understanding. It leverages LVLMs for semantic goal inference but still relies on a relatively tightly coupled architecture between perception, language reasoning, and control, limiting flexibility in real-world deployment.

More recently, the trend has shifted toward using large vision-language models as frozen backbones with lightweight adaptation. NaVILA~\cite{navila2023} is an example of this direction, utilizing frozen CLIP models to enable instruction-conditioned navigation without heavy fine-tuning. By distilling navigation goals into vision-language-aligned spaces, NaVILA reduces the need for extensive retraining while achieving strong generalization across tasks.

MapNav\cite{mapnav2024} takes a complementary approach by constructing annotated semantic maps from vision-language inputs, achieving state-of-the-art navigation results. However, its reliance on a specialized map encoder and structured annotations limits its flexibility.
Mem2Ego\cite{mem2ego2024} further pushes the boundary by introducing a global-to-ego memory mechanism for dynamic long-horizon planning, but at the cost of requiring fine-tuning of large vision-language models and primarily being validated on synthetic datasets such as HSSD.

In addition, approaches like VLMaps\cite{vlmaps2023} and EVA-Nav\cite{evanav2023} propose building semantic representations or embodied policies directly from pre-trained LVLMs, often freezing most parameters to enhance modularity and transferability.

In contrast to these prior works, our project proposes a plug-and-play modular framework that cleanly decouples vision-language understanding and navigation policy. By treating semantic understanding and action planning as independent but interoperable modules, we leverage the rich representational power of large vision-language models while maintaining fast, efficient inference suitable for real-world deployment. Our design philosophy emphasizes minimal fine-tuning, high adaptability, and practical scalability across different environments and tasks.
\section{Methods}
\label{sec:methods}
\begin{figure*}[t]
    \centering
    \includegraphics[width=\textwidth]{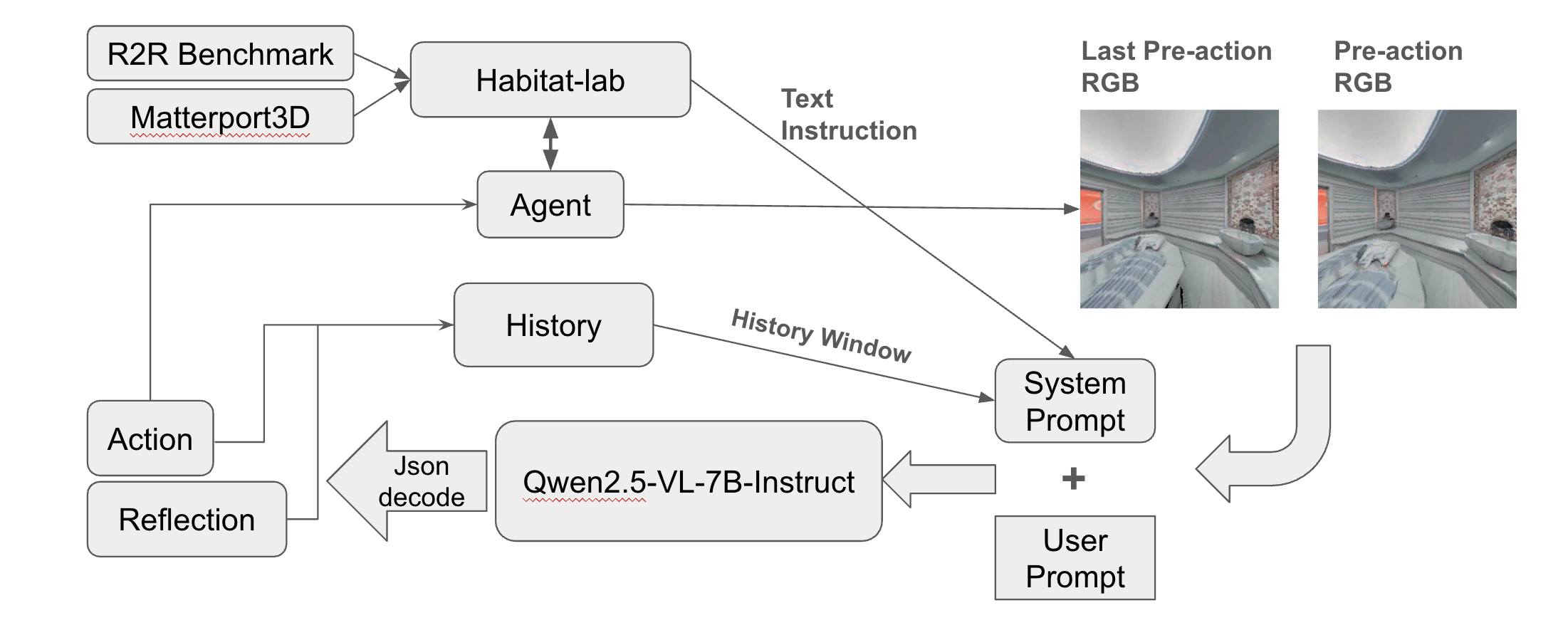}
    \caption{Overview of our framework. The system processes vision and language input jointly, manages history context, and iteratively predicts actions to complete navigation tasks.}
    \label{fig:framework_overview}
\end{figure*}

Our framework is generally described in Figure~\ref{fig:framework_overview}.

\subsection{Dataset}
We use the Matterport3D (MP3D) dataset as the 3D scene source for our navigation framework. MP3D provides high-resolution RGB-D scans of large-scale indoor environments, enabling realistic geometric and semantic context for agent navigation and visual-language grounding tasks \cite{DBLP:journals/corr/abs-1709-06158}.

\subsection{Vision-Language Model}
We employ the Qwen2.5-VL-7B-Instruct model as the core component of our vision-and-language navigation (VLN) framework. This multimodal large language model is adept at interpreting visual inputs and natural language instructions, subsequently generating actionable outputs for agent navigation. In particular, Qwen2.5-VL-7B-Instruct has demonstrated strong performance on various agent benchmarks, including achieving 84.7\% on ScreenSpot, 81.9\% on AITZ\_EM, and 91.4\% on MobileMiniWob++\_SR, underscoring its capability in visual understanding and decision-making tasks \cite{qwen2.5-VL, Qwen2VL}.

\subsection{Simulation Environment}
We utilize Habitat-Lab as our simulation and benchmarking environment. Habitat-Lab is a modular high-level library designed for training and evaluating embodied AI agents across a variety of tasks and environments. It provides flexible task definitions, diverse embodied agents, and tools for benchmarking performance using standard metrics \cite{puig2023habitat3,szot2021habitat}. By integrating the Matterport3D dataset into Habitat-Lab, we perform our navigation tasks in a continuous environment setting, aligning with the VLN-CE framework. This approach contrasts with the original Matterport3D-based VLN tasks, which operated in discrete environments with predefined navigation graphs, thereby offering a more realistic and unconstrained navigation experience \cite{krantz_vlnce_2020}.

\subsection{Prompt Engineering Strategy}
\subsubsection{History Management}
\label{sec:History Management}
We maintain a structured history buffer \( \mathcal{H} \) to provide bidirectional context for the Vision-Language Model (VLM) agent. Each history entry at time step \( t \) is represented as a tuple:
\[
h_t = (\text{step}_t, \text{action}_t, \text{reflection}_t)
\]
where \(\text{step}_t\) denotes the step count, \(\text{action}_t\) is the action taken, and \(\text{reflection}_t\) is the agent's self-generated reflection. During each retrieval, a fixed-size window parameter \( W \) determines the maximum number of past entries \(\{h_{t-W}, \ldots, h_{t-1}\}\) included in the agent's context.

\subsubsection{Prompt Design and Visual Input Strategy}

We adopt a standard structure consisting of a system prompt and a user prompt for Vision-Language Model (VLM) inference in agentic tasks. The system prompt is organized into the following components:
\begin{enumerate}
    \item \textbf{Persona}: Describes the agent's character and capabilities.
    \item \textbf{Agent Parameters}: Specifies navigation parameters such as turning angle, movement distance, and available action space. Following the R2R benchmark definition~\cite{krantz_vlnce_2020}, the agent operates with four discrete actions:
\[
\{\texttt{turn\_left}, \texttt{turn\_right}, \texttt{move\_forward}, \texttt{stop}\}.
\]
    \item \textbf{Human Common Sense}: Provides general principles of human indoor navigation behavior.
    \item \textbf{History Context}: Encodes past interactions \(\{h_{t-W}, \ldots, h_{t-1}\}\) as defined in Section~\ref{sec:History Management}.
\end{enumerate}

The user prompt follows a standard video-based inference format. Its primary purposes are:
\begin{enumerate}
    \item \textbf{Reflection Reactivation}: Reinforces the agent's prior reasoning and situational awareness.
    \item \textbf{Output Formatting}: Ensures the output follows a structured JSON schema, consisting of one predicted action and one generated reflection, to facilitate downstream parsing and execution.
\end{enumerate}

For visual input, we employ a two-frame strategy, providing the model with:
\[
\mathcal{V} = \{\mathbf{I}_{t-1}, \mathbf{I}_{t}\}
\]
where \(\mathbf{I}_{t-1}\) is the RGB image captured before the last action, and \(\mathbf{I}_{t}\) is the current RGB image after executing the latest action. This design offers temporal continuity and enhances visual grounding across actions. Implementation details can be found at our GitHub (Section~\ref{sec:github}).

\subsection{Inference Loop Structure}

Our inference process forms a loop-based structure, facilitated by the integration of Habitat-Lab. At each time step, the agent processes the visual and language input to predict the next action. The loop continues until one of the stopping criteria is met: reaching the maximum number of steps (\texttt{max\_step}), achieving the navigation goal (success), or the agent issuing a self-instructed \texttt{stop} action. This design ensures that the agent can autonomously decide when to terminate navigation, balancing between exploration and task completion.

\section{Experiments}
\label{sec:experiments}

We evaluated our framework on the Room-to-Room (R2R) benchmark within the Vision-and-Language Navigation in Continuous Environments (VLN-CE) setting. Specifically, we focus on the first 20 trajectories of the validation unseen (val-unseen) split \cite{krantz_vlnce_2020}. Our simulation settings follow the original VLN-CE Room-to-Room configuration, with two key modifications: we limit the maximum number of navigation steps to 50 and increase the agent's camera resolution to $256 \times 256$ pixels. These adjustments impose stricter evaluation conditions and provide higher-quality visual inputs for our framework.

Our evaluation metrics include the following.

\begin{itemize}
  \item \textbf{Distance to Goal (DTG)}: The average Euclidean distance (in meters) between the agent's final position and the goal location.
  \item \textbf{Success Rate (SR)}: The percentage of episodes in which the agent ends within 3 meters of the goal.
  \item \textbf{Success weighted by Path Length (SPL)}: A metric that accounts for both the success rate and the efficiency of the path taken, defined as:
  \[
  \text{SPL} = \frac{1}{N} \sum_{i=1}^{N} S_i \cdot \frac{l_i}{\max(p_i, l_i)}
  \]
  where \( S_i \) is a binary indicator of success for episode \( i \), \( l_i \) is the shortest path length and \( p_i \) is the actual path length taken by the agent.
\end{itemize}

We compare our results with two baseline methods: BEVBert \cite{an2023bevbert} and ETPNav \cite{an2024etpnav}. The performance metrics for the baselines are summarized in Table~\ref{tab:baselines}, and our results are presented separately in Table~\ref{tab:ours} due to the limited evaluation on only 20 trajectories.

\begin{table}[h]
\centering
\caption{Performance of baseline methods on the R2R VLN-CE val-unseen split.}
\label{tab:baselines}
\begin{tabular}{lccc}
\toprule
\textbf{Method} & \textbf{DTG (m)} & \textbf{SR (\%)} & \textbf{SPL (\%)} \\
\midrule
BEVBert & 2.81 & 75.0 & 64.0 \\
ETPNav & 3.95 & 66.0 & 59.0 \\
\bottomrule
\end{tabular}
\end{table}

\begin{table}[h]
\centering
\caption{Our framework's performance on the first 20 trajectories of the R2R VLN-CE val-unseen split.}
\label{tab:ours}
\begin{tabular}{lccc}
\toprule
\textbf{Method} & \textbf{DTG (m)} & \textbf{SR (\%)} & \textbf{SPL (\%)} \\
\midrule
\textbf{Zero movement} & 8.150 & 0.0 & 0.0 \\
\textbf{Ours} & \textbf{7.748} & 5.0 & 5.0 \\
\bottomrule
\end{tabular}
\end{table}

Our results indicate that, under the stricter evaluation conditions - specifically, a reduced maximum of 50 steps - our framework did not achieve successful navigation within the first 20 trajectories of the R2R VLN-CE val-unseen split. The agent's average Distance to Goal (DTG) was 7.748 meters, with both Success Rate (SR) and Success weighted by Path Length (SPL) at 5.0{\%}. These outcomes suggest that our current framework struggles to generalize effectively in unseen environments under these more challenging conditions.

\subsection{Discussion and Future Work}

Future work will address the current challenges by expanding evaluation, fine-tuning the model, and exploring the integration of additional sensory inputs, such as simulating a topological graph based on current view geometries. Although our goal was to design a navigation framework independent of explicit navigation graphs, our results suggest that incorporating structured priors may still be critical for achieving robust performance in complex environments.
\subsubsection*{Code Availability}
\label{sec:github}
The code and additional resources for this project are publicly available at \url{https://github.com/YichengDuan/oobvlm}.

{
    \small
    \bibliographystyle{ieeenat_fullname}
    \bibliography{main,reference}
}
% WARNING: do not forget to delete the supplementary pages from your submission 
% \input{sec/X_suppl}

\end{document}